\documentclass[letterpaper, 10 pt, conference]{ieeeconf}  

\IEEEoverridecommandlockouts                              

\usepackage{amsmath} 
\usepackage{amssymb}  
\usepackage[dvipsnames]{xcolor}
\usepackage{booktabs}
\usepackage{multirow}
\usepackage{bm}
\usepackage{graphicx}
\usepackage{mathtools}
\usepackage{float}
\usepackage{subfigure}
\usepackage{overpic}
\usepackage{lipsum}
\usepackage{tabularx}

\title{\LARGE \bf
3DGS-Calib: 3D Gaussian Splatting for Multimodal\\ SpatioTemporal Calibration
}

\author{Quentin Herau$^{1,3}$,
Moussab Bennehar$^{1}$,
Arthur Moreau$^{2}$,
Nathan Piasco$^{1}$,
Luis Roldão$^{1}$,
Dzmitry Tsishkou$^{1}$,\\
Cyrille Migniot$^{4}$,
Pascal Vasseur$^{5}$
and Cédric Demonceaux$^{3}$
\thanks{$^{1}$ Noah's Ark, Huawei Paris Research Center, France. {\tt\small \{Quentin.Herau, Nathan.Piasco, Moussab.Bennehar, Luis.Roldao, Dzmitry.Tsishkou\}@huawei.com}}
\thanks{$^{2}$ Noah's Ark, Huawei London Research Center, United Kingdom. {\tt\small Arthur.Moreau2@huawei.com}}
\thanks{$^{3}$ ICB UMR CNRS 6303, Universit\'{e} de Bourgogne, France. {\tt\small \{Quentin.Herau@etu., Cedric.Demonceaux@\}u-bourgogne.fr }}
\thanks{$^{4}$ ImViA UR 7535, Universit\'{e} de Bourgogne, France. {\tt\small  Cyrille.Migniot@u-bourgogne.fr }}
\thanks{$^{5}$ MIS UR 4290, Universit\'{e} de Picardie Jules Verne, France. {\tt\small Pascal.Vasseur@u-picardie.fr}}
}

\begin{document}
\maketitle
\thispagestyle{empty}
\pagestyle{empty}


\begin{abstract}
Reliable multimodal sensor fusion algorithms require accurate spatiotemporal calibration. Recently, targetless calibration techniques based on implicit neural representations have proven to provide precise and robust results. Nevertheless, such methods are inherently slow to train given the high computational overhead caused by the large number of sampled points required for volume rendering. 
With the recent introduction of 3D Gaussian Splatting as a faster alternative to implicit representation methods, we propose to leverage this new rendering approach to achieve faster multi-sensor calibration. We introduce 3DGS-Calib, a new calibration method that relies on the speed and rendering accuracy of 3D Gaussian Splatting to achieve multimodal spatiotemporal calibration that is accurate, robust, and with a substantial speed-up compared to methods relying on implicit neural representations. We demonstrate the superiority of our proposal with experimental results on sequences from KITTI-360, a widely used driving dataset.
\end{abstract}

\section{INTRODUCTION}
\label{sec:introduction}
Sensors are essential in robotics and intelligent systems when it comes to providing knowledge about the surrounding environment. Depending on the type of the sensor, different information modalities are provided (e.g. geometric range information with LiDAR and color information with RGB cameras).
Combining all information from multiple sensors allows scene understanding to be maximized, enabling improved performance in crucial tasks such as localization, mapping, and object detection. 
However, to correctly merge and exploit the data coming from the various sensors, it is necessary to first perform an accurate spatial and temporal calibration phase.

The classical strategy of performing spatial calibration is through the use of one or more targets of known characteristics carefully positioned around the scene, such as checkerboards~\cite{geiger2012automatic} or other custom targets~\cite{guindel2017automatic,pusztai2017accurate}. The goal of these methods is to have multiple acquisitions of the target with both sensors (e.g. LiDAR depth sensor and RGB cameras) and to find matching information within the two modalities (plane detection in the LiDAR point cloud and corner detection in the camera RGB images). By finding correspondences, these methods can determine the correct spatial calibration.
Classical methods are accurate and robust, but hardly scalable as they require manual placement of targets in addition to structural assumptions about the surrounding environment. Such requirements are incompatible with in-the-wild re-calibration where targets may not be available, or placing them might be impractical.

To remove these constraints, targetless calibration methods directly exploit existing information within the scene without requiring specific user-placed targets.
These methods can either rely on specific features naturally present in the scene (e.g. edges~\cite{napier2013cross,yuan2021pixel} or intensity~\cite{taylor2012mutual,pandey2012automatic}) or extract features automatically by relying on neural networks~\cite{iyer2018calibnet,lv2021lccnet}.
Although these methods do not require manual user-placed targets, they still depend on the presence of either specific features in the scene or are supervised using data provided by a sensor setup with known calibration parameters.

Recently, a new family of calibration methods based on implicit neural representations (i.e. NeRF: Neural Radiance Fields) has emerged~\cite{zhou2023inf,herau2023moisst,herau2023soac} demonstrating impressive calibration results without requiring as many priors as classical methods. NeRF-based calibration strategies build a geometrically consistent and continuous volumetric representation of the scene enforcing consistency between all the sensors. Thus, these approaches achieve robust and accurate calibration without the limitations of the previously mentioned methods. Nevertheless, the extended training times associated with a NeRF model create a significant limitation hindering their adoption in real-world applications.

Recently, 3D Gaussian Splatting (3DGS)~\cite{kerbl20233d} was introduced as a novel 3D scene representation with significantly faster training times than NeRF and real-time rendering, while providing on-par rendering quality. Thanks to its advantages compared NeRF, 3D Gaussian Splatting has gained increasing popularity and has been applied to solve different tasks such as rendering human avatars~\cite{moreau2023human} and reconstructing in-the-wild~\cite{hiba2024swag} or driving scenes~\cite{zhou2023drivinggaussian}.

Motivated by their advantages and given their shared properties with NeRF, we present in this paper  \textbf{3DGS-Calib}, a novel targetless multimodal spatiotemporal calibration solution that capitalizes on the speed and quality of 3D Gaussian Splatting. By exploiting the LiDAR point cloud as a reference for the Gaussians' positions, we learn a continuous representation of the scene. Using this representation and enforcing both geometrical and photometric consistency between all the sensors enables us to achieve accurate calibration requiring less training time compared to NeRF-based methods.
Our method offers accurate and robust calibration and significantly outpaces existing NeRF-based methods. 

\noindent In summary, our main contributions are as follows:
\begin{itemize}
    \item To the best of our knowledge, we propose the first multimodal spatiotemporal calibration method based on 3DGS~\cite{kerbl20233d}, which allows significant speed-up while providing better accuracy and robustness than NeRF-based methods,
    \item We evaluate our method on sequences from KITTI-360~\cite{liao2022kitti}, a widely used driving dataset, and show that our method provides better results on driving scenes for LiDAR/Camera calibration, without the need for any supervision or specific features in the scene.
\end{itemize}


\section{RELATED WORK}
\label{sec:related_work}
\noindent\textbf{Targetless Calibration.}
Classical targetless multimodal calibration methods rely on specific features present in the scene that can be matched between the different modality sensors (e.g. LiDARs and RGB cameras) to achieve the calibration goal. For instance, Taylor et al.~\cite{taylor2012mutual} and Pandey et al.~\cite{pandey2012automatic} propose to apply a correspondence between the grayscale level of the color images and the intensity of the LiDAR scans. 
Nevertheless, such a correlation is not always correct (e.g. in the case of shadows within the scene).
On the other hand, Napier et al.~\cite{napier2013cross} and Yuan et al.~\cite{yuan2021pixel} match edges detected in both the images and the LiDAR scans, to align the scene structures extracted from the different modalities. However, point cloud-based edge detection degrades with the sparsity of the acquisition, requiring the sensors to stay still during the process to gather more data or needing more expensive systems.

Other lines of work like CalibNet~\cite{iyer2018calibnet} and LCCNet~\cite{lv2021lccnet} rely on the ability of neural networks to automatically extract high-level features for calibration with deep learning models.
Such methods can predict the correct calibration between LiDAR scan and image pairs in real-time, enabling online re-calibration. Nevertheless, a labeled dataset is needed to train the models, creating a bias towards the type of scenes used for training. Furthermore, as shown in~\cite{herau2023soac}, these methods are setup-specific and need to be re-trained if there is a considerable difference with the training data.

More recently, Neural Radiance Fields (NeRF)~\cite{mildenhall2021nerf} were introduced enabling an implicit representation of a 3D scene learned within a neural network. Seminal works have exploited this representation to solve a wide range of applications such as novel view synthesis \cite{barron2022mip,muller2022instant}, localization~\cite{moreau2023crossfire}, surface reconstruction~\cite{wang2021neus} or pose registration~\cite{wang2021nerf,lin2021barf}. Notably, some recent works propose to rely on NeRF for multimodal sensor calibration~\cite{zhou2023inf,herau2023moisst,herau2023soac}. This is done by exploiting the fully differentiable structure of the NeRF model to simultaneously learn a volumetric scene representation while regressing the correct calibration parameters to optimize the rendering quality.
INF~\cite{zhou2023inf} proposes to first train the density of the scene with the LiDAR scans, before calibrating the extrinsic parameters of a 360° camera by learning the color information from the captured images. MOISST~\cite{herau2023moisst} performs spatial and temporal calibration of multiple LiDARs and cameras by building a trajectory with the poses of a reference sensor and training a NeRF with the information provided by all the sensors. SOAC~\cite{herau2023soac}, while requiring a camera as the reference sensor, further improves the robustness and accuracy by using multiple NeRFs and taking into account the overlap between the sensors over the sequence. While these methods can provide high accuracy and robustness, they require relatively long training times (1-2 hours~\cite{herau2023soac}) even when exploiting more efficient hash-based encoding techniques~\cite{muller2022instant}.  
\newline

\noindent\textbf{3D Gaussian Splatting.}
3D Gaussian splatting~\cite{kerbl20233d} is a novel volumetric rendering method that exploits the high efficiency of splatting 3D Gaussians on a 2D image to obtain real-time rendering of novel views. Unlike NeRF, 3DGS features an explicit representation composed of a high number of Gaussians each defined by its position, color, scale, and opacity. These parameters are trained with gradient descent by applying a loss function between the rendered image and its ground-truth counterpart provided by the camera to minimize the photometric error.
Since it is based on an explicit representation, 3DGS requires an initial point cloud to define the Gaussians at the beginning of the training, which is commonly obtained through Structure-from-Motion.
During training, the model is allowed to add new Gaussians to create more details or prune them if deemed useless for the representation.

As the rasterization of 3D Gaussian splatting is differentiable with respect to the positions of the Gaussians, some methods propose to propagate the gradients to the image poses to solve tasks like SLAM~\cite{yan2023gs,matsuki2023Gaussian} or image registration~\cite{fu2023colmap}. These methods use the depth information from RGB-D cameras~\cite{yan2023gs,matsuki2023Gaussian}, or the predicted monocular depth~\cite{fu2023colmap} to create a point cloud for the initialization of the Gaussians. In these methods, during the pose optimization/localization step, the model minimizes the photometric cost function by optimizing the pose of each newly introduced image of the sequence with respect to the already existing parameterized Gaussians.
The number of Gaussians and their parameters in these approaches are fixed to prevent the model from compensating the incorrect pose with a degenerate scene representation. Once a new image has been registered, it can be used with its obtained camera pose to train the representation and eventually add new Gaussians. 
This strategy can also be transposed to the calibration of multimodal systems which, as far as we know, has not yet been addressed. Thus, we have decided to tackle this subject in this work.

\section{METHOD}
\label{sec:method}
\begin{table}[hbt!]
\centering
\vspace{-2mm}
\caption{\label{tab:notations} Notations used in this paper.
}
\scriptsize
\setlength{\tabcolsep}{0.01\linewidth}
\renewcommand{\arraystretch}{1.15}
\vspace{-3mm}
\begin{tabular}{c | l }
    Notation & Meaning \\
    \midrule
    $C$ & Set of cameras\\
    $\{F_i\}_{i\in C}$ & Sets of frames captured by the camera $i\in C$\\
    $t^{n_i} \in \mathbb{R^+} $ & Timestamp of images $n_i\in F_i$ relative to the camera $i\in C$\\
    $\delta_i \in \mathbb{R}$ & Time offset between the LiDAR and the camera $i\in C$\\
    $\prescript{}{w}{T}^i(t)\in \mathbb{R}^{4\times 4}$ & The pose of camera $i\in C$ at time $t$ (the time is relative to \\
    &sensor $i$'s own clock) in the world reference frame\\
    $\prescript{}{w}{T}^L(t)\in \mathbb{R}^{4\times 4}$ & Pose of the LiDAR at time $t$\\
    $\prescript{}{L}{T}^i \in \mathbb{R}^{4 \times 4}$ & Transformation matrix from camera $i$ to LiDAR\\
    $X \in \mathbb{R}^{3 \times G}$ & Accumulated LiDAR point positions in the world reference,\\ &
    with $G$ the number of points\\
\bottomrule
\end{tabular}
\vspace{-4mm}
\end{table}
\begin{figure*}
    \centering
    \includegraphics[width=0.8\textwidth]{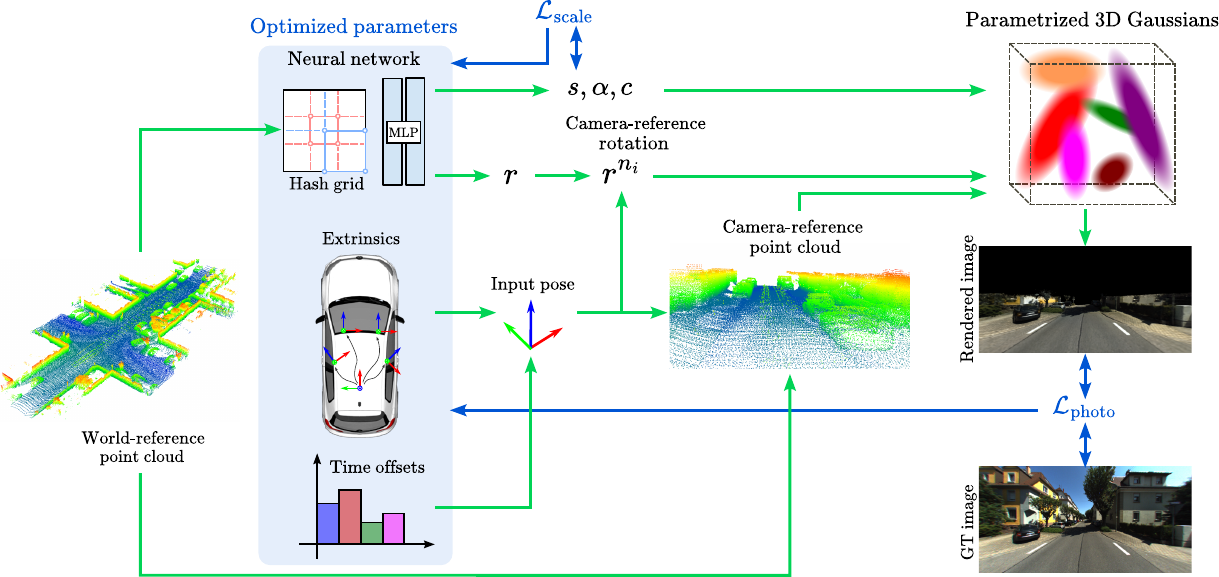}
    \caption{\textbf{Pipeline of 3DGS-Calib:} The Gaussians' positions are given as input to the neural network which predicts their parameters. In parallel, the calibration parameters provide the input pose that transforms the Gaussians from the world frame to the image frame. Then, the 3D Gaussians are splatted using their predicted parameters to generate the rendered image. This image is compared to its ground-truth (GT) counterpart to compute the photometric loss. Finally, the gradients are backpropagated to the neural network and the calibration parameters.}
    \label{fig:method}
\end{figure*}

In this paper, we formulate the multi-modal calibration problem as follows. Given the poses of the LiDAR, which can be derived either through reliance on LiDAR-based SLAM~\cite{lego_loam} or simply via the Iterative Closest Point (ICP) algorithm~\cite{arun1987least,vizzo2023kiss}, in addition to initial noisy/inaccurate priors of the spatial and temporal calibration parameters w.r.t. the LiDAR, our goal is to obtain the correct calibration parameters. In contrast to NeRF-based methods where one of the cameras is often considered as the reference sensor~\cite{herau2023moisst}, our method uses the LiDAR as its reference sensor since we rely on its point cloud to initialize the 3D Gaussians' positions. Our goal is to obtain the correct spatiotemporal calibration of the cameras on the vehicle w.r.t. the LiDAR. The overall pipeline is described in Figure~\ref{fig:method}.

\subsection{Notations and Background} \label{sec/method/notations}

The notations used throughout this paper are summarized in Table~\ref{tab:notations}.
Our objective is to determine the optimal spatial transformations $\hat{\prescript{}{L}{T}^i}$ and time offsets $\hat{\delta_i}$ w.r.t. the LiDAR for the various sensors to achieve calibration. Drawing from similar methodologies employed in MOISST~\cite{herau2023moisst} and SOAC~\cite{herau2023soac}, we establish a continuous trajectory, denoted as $\mathcal{T}_L$, for the LiDAR from its discrete poses. Linear interpolation is used for pose translation, while spherical linear interpolation (SLERP~\cite{shoemake1985animating}) is employed for rotation. This trajectory is represented as a time-dependent function, $\prescript{}{w}{T}^L(t) = \mathcal{T}_L(t)$ which provides the LiDAR pose for any given time $t$.
By leveraging the extrinsic transformations and time offsets between the cameras and the LiDAR, we are able to calculate the absolute pose of camera $i$ at particular timestamps using the following equation:
\begin{equation}
     \prescript{}{w}{T}^i(t^{n_i}+\delta_{i}) =\mathcal{T}_L(t^{n_i}+\delta_{i}) \prescript{}{L}{T}^i.
     \label{eq:ext}
\end{equation}
From now on, we designate the absolute pose of camera $i$ computed from its extrinsic as $T^{n_i} = \prescript{}{w}{T}^i(t^{n_i}+\delta_{i})$.

\subsection{Scene representation with 3D Gaussians}
3DGS~\cite{kerbl20233d} models the scene using parameterized 3D Gaussians, which is an explicit representation, unlike NeRF~\cite{mildenhall2021nerf}. The Gaussians are parameterized by their centers $\mu \in \mathbb{R}^{3 \times G}$, their opacities $\alpha \in \mathbb{R}^G$, their rotation quaternions $r \in \mathbb{R}^{4 \times G}$, their 3D scales $s \in \mathbb{R}^{3 \times G}$ and their RGB colors $c \in \mathbb{R}^{3 \times G}$. In our case, as we focus on calibration and not on rendering quality, we consider the colors as view-independent. This helps in avoiding degenerate cases where color changes from different viewing directions can compensate for noisy poses.
As we have access to the LiDAR scans and their associated poses, it means that we can accumulate these scans to obtain a point cloud of the sequence. This point cloud $X$ is used to define the positions of the Gaussians. As these points are supposed to represent a real 3D point seen by the LiDAR, we do not optimize their positions, which we consider accurate from the beginning. Thus, $\mu$ parameters are not needed in our case as we replace them using $X$. While 3DGS learns the scene geometry (i.e. Gaussians' positions) from known camera poses, our key idea is to register the camera poses on the scene geometry from the LiDAR. In the same way as the 3DGS SLAM methods~\cite{yan2023gs,matsuki2023Gaussian}, we also remove densification and pruning, to prevent the model from creating or removing Gaussians which would allow the model to overfit to each image with incorrect geometry. 
For each training step, by using the calibration parameters to determine the input pose, we transform the Gaussians to the image frame:
\begin{equation}
    X^{n_i} = (T^{n_i})^{-1}X,
\end{equation}
with $X^{n_i}$ being the Gaussians' centers in the camera frame associated with image $n_i$. Similarly, the anisotropic Gaussians are rotated using:
\begin{equation}
    r^{n_i} = (R^{n_i})^{-1}r,
\end{equation}
with $R^{n_i}$ being the rotation matrix extracted from $T^{n_i}$ and $r^{n_i}$ the Gaussian rotations in the camera frame associated with image $n_i$.
Then, the 3D Gaussians are splatted to render an image according to the intrinsic parameters of the camera. The same photometric loss $\mathcal{L}_{photo}$ function as the original 3DGS paper~\cite{kerbl20233d} is used:
\begin{equation}
    \mathcal{L}_{\text{photo}} = (1-\lambda_1) \mathcal{L}_1 + \lambda_1 \mathcal{L}_{\text{D-SSIM}},
\end{equation}
with $\lambda_1 = 0.2$ in our experiments.
The goal is to find the optimal parameters such as:
\begin{equation} \label{eq:opti}
\begin{multlined}
        \left\{ \hat{\prescript{}{L}{T}^{i}}, \hat{\delta}_i \right\}_{i \in C}, \hat{\alpha}, \hat{s}, \hat{r}, \hat{c}= \\
        \underset{\left\{ \prescript{}{L}{T}^{i}, \delta_{i},  \alpha, s, r, c \right\}}{\text{argmin}}(\mathcal{L}_{\text{photo}}(\Phi(X^{n_i},\alpha,s,r^{n_i},c),n_i))
\end{multlined}
\end{equation}
with $\Phi$ the rendering function from~\cite{zwicker2001ewa}.
The gradients are then backpropagated to both the Gaussian parameters, and the calibration parameters through the input pose.

\subsection{Gaussian parameters with Neural Network}

Learning independent Gaussians' parameters while modifying the pose of several sensors leads to a suboptimal training process. This is mainly because every change in the calibration parameters will require each Gaussian to adapt its information about the scene separately.
To address this issue, we learn the Gaussians' parameters using a neural network that provides a regularized scene representation shared across sensors.
We use the model from Instant-NGP~\cite{muller2022instant}, which takes as input the 3D position of a Gaussian. We add 4 heads at the output of the Multilayer Perceptron (MLP) to predict the color, opacity, scale, and rotation parameters of the Gaussian.
This allows the spatially close Gaussians to have similar parameters, which is coherent with the nature of real-world scenes where the colors and shapes are continuous on large areas (ex: walls, roads).
Thus, we define the function $F(X|\Theta)$ which takes as input the 3D positions of the Gaussian, and returns the Gaussian parameters according to the network parameters $\Theta$.
We then redefine Eq.~\ref{eq:opti} as follows:
\begin{equation} \label{eq:opti_mlp}
\begin{multlined}
        \left\{ \hat{\prescript{}{L}{T}^{i}}, \hat{\delta}_i \right\}_{i \in C},\hat{\Theta}= \\
        \underset{\left\{ \prescript{}{L}{T}^{i}, \delta_{i}, \Theta \right\}}{\text{argmin}}(\mathcal{L}_{\text{photo}}(\Phi(X^{n_i},F(X|\Theta)),n_i))
\end{multlined}
\end{equation}
As the Gaussians' positions are transformed to the camera frame, we remove the points behind the camera before inference, allowing a considerable time speed-up. 

\subsection{Preprocessing and optimization}
To improve both the performance and the training speed of our method, we adopt the following enhancing design decisions:
\subsubsection{Accumulated LiDAR downsampling}
When accumulating the LiDAR scans, as multiple passes are done in the same area over time, this results in a large number of redundant points, which not only does not improve the rendering quality but also slows down the training process, as each Gaussian requires an MLP inference. By using voxel-based downsampling, we can obtain a homogeneously dense point cloud at the desired resolution. This solution allows considerable speed-up of the calibration process. As we keep one point per voxel, we also enforce the scale of each Gaussian to be smaller than the voxel size. This helps the rendered result to be sharper, by preventing the Gaussians from overlapping each other.


\subsubsection{Coarse-to-fine voxelization}
Since the Gaussians' parameters are obtained through the neural network inference, the training time is proportional to the number of the used points. During the beginning of the calibration process, as we are still correcting large miscalibration errors, it is not necessary to have a high level of detail in the rendered images. Thus, we can use larger voxels for the downsampling. As the training progresses, we increase the resolution since we are getting more precise calibration. This allows the first part of the training to run much faster.

\subsubsection{Image cropping}
The Gaussians' positions in our method are exclusively determined by the LiDAR points which, in typical driving sensor setups, mostly cover the road and the lower/closer structures of the environment. The LiDAR points are unable to provide the structure of far elements and the sky. In light of this, we only consider the bottom half of the images when calculating the losses.

\subsubsection{Gaussians' scale regularization}
During the training, to avoid very thin Gaussians, which usually create artifacts in the representation, we apply a scale regularization loss similar to~\cite{matsuki2023Gaussian}, which prevents a high scale difference along the 3 axes.
This is achieved using the following regularization loss:
\begin{equation}
    \mathcal{L}_{\text{scale}} = \sum_{g=1}^{G} \| s_g - \tilde{s}_g \|_1,
\end{equation}
with $G$ being the total number of the Gaussians, and $\tilde{s}$ the mean of the scale along the 3 axis.
Hence, our model's loss function becomes:
\begin{equation}
    \mathcal{L}_{\text{total}} = \mathcal{L}_{\text{photo}} + \lambda_2 \mathcal{L}_{scale},
\end{equation}
with $\lambda_2 = 0.001$ across all our tests.

\section{EXPERIMENTS}
\label{sec:experiments}
\begin{figure*}[!htbp]
\centering
\setlength{\tabcolsep}{0.001\linewidth}
    \begin{tabular}{ccc}
         \includegraphics[height=.28\linewidth]{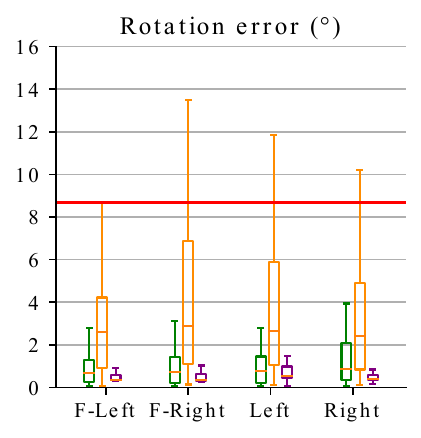}& \includegraphics[height=.28\linewidth]{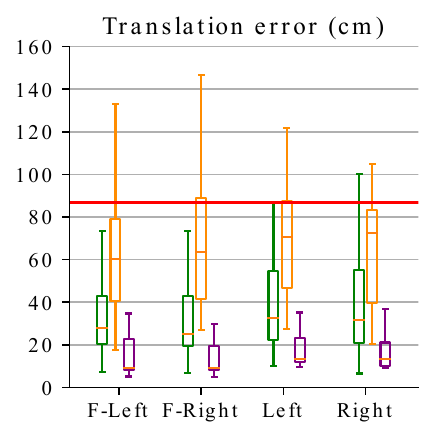}&
         \includegraphics[height=.28\linewidth]{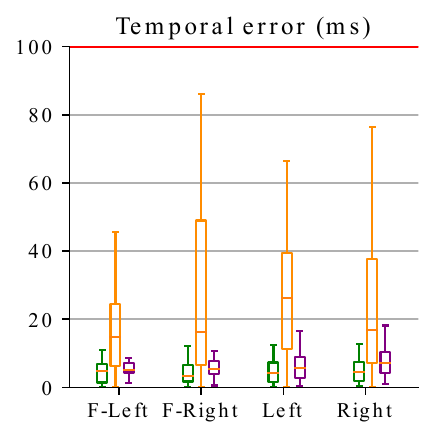}
    \end{tabular}
\caption{Results for \textcolor{ForestGreen}{MOISST}, \textcolor{orange}{MOISST /w cropping} and \textcolor{violet}{3DGS-Calib} as box plots. The \textcolor{red}{red lines} show the initial error (Best viewed in color).}
\label{fig:results}
\vspace{-5mm}
\end{figure*}
\subsection{Dataset}
For our evaluation, we use the KITTI-360 driving dataset~\cite{liao2022kitti}, which contains images from 2 front cameras and 2 side fish-eye cameras in addition to point clouds from a 360° LiDAR. We build 3 sequences with different characteristics to evaluate our method in different conditions. The built sequences are described in Table~\ref{tab:kitti_sequences}. We skip 1 out of 2 frames, which results in 40 frames for each sensor in every sequence.
\begin{table}[hbt!]
\centering
\scriptsize
\caption{\label{tab:kitti_sequences} Selected frames for each KITTI-360 sequence.}

\renewcommand{\arraystretch}{1.2}
\begin{tabular}{@{}c c c c c @{}}
Sequence & Run & Starting frame & Ending frame & Description\\
\toprule
1 & 0009 & 980 & 1058 & Straight line\\
2 & 0009 & 2854 & 2932 & Large rotation\\
3 & 0010 & 3390 & 3468 & Small zigzag\\
\bottomrule
\end{tabular}
\vspace{-5mm}
\end{table}
\subsection{Training details}
Our model is trained for a total number of 6000 steps. We initialize Gaussians' positions using a downsampled point cloud with a voxel size of 10 cm for 4000 steps.
Then, we reduce the downsampling voxel size to 5 cm and train for 1000 steps. Finally, for the finest rendering details and calibration parameters, we use a voxel size of 2 cm for 1000 steps.
To stabilize the training, we warm up for 500 steps before kicking off the calibration process. The colors are randomly initialized, and the scales, rotations and opacities are initialized as in the original 3DGS implementation.\\ 
As proposed in MOISST~\cite{herau2023moisst} and SOAC~\cite{herau2023soac}, a weight decay of $10^{-4}$ is applied on the hash grid to allow better convergence during the whole training. 
We use the Adam optimizer and apply a learning rate of $10^{-4}$ for rotation, $5\times10^{-3}$ for translation, and $10^{-3}$ for temporal correction. These learning rates are linearly decayed by 0.1 times the initial value at the end of the training.

When comparing with MOISST, we use the same parameters as the original paper, except that we limit the number of epochs to 10, as we observed that the model convergences. We also start the depth loss from the beginning of the training, as the LiDAR is assumed to have correct poses, and thus provide correct geometry.
For fairness, we compare our method with MOISST using both the whole image and its bottom half as it is done in our method. The reasoning behind this is to provide both methods with the same prior and knowledge about the sensor setup.
For the initial noise, we set an error of 5° and 50~cm on all 3 axes, which can be either positive or negative. In the same way, we add a 100~ms time offset on the sensors to simulate synchronization errors. All experiments are run with 10 random seeds before being averaged.
We run all methods with the same GPU whose performance is equivalent to an NVIDIA RTX 3090.

\subsection{Spatiotemporal calibration}
For spatiotemporal calibration, we compare our method to MOISST, as it can use the LiDAR as the reference sensor (in contrast to SOAC~\cite{herau2023soac}), and provide both spatial and temporal calibration. In Figure~\ref{fig:results}, we provide the results of our method in boxplots\footnote{The boxes show the first quartile $Q_1$, median, and third quartile $Q_3$. The whiskers use 1.5 IQR (Interquartile range) above and below the box and stop at a value within the results.} against MOISST trained with the whole image or only the bottom half of the images (w/ cropping).
\begin{table}[hbt!]
\centering
\scriptsize
\caption{\label{tab:training_time} Training time for each sequence in seconds.}

\renewcommand{\arraystretch}{1.2}
\begin{tabular}{@{}c c c c c c @{}}
Sequence & MOISST & MOISST w/ cropping & 3DGS-Calib  \\
\toprule
1 & $5340~(\times10.39)$ & $2707~(\times5.25)$ & $\mathbf{514}$ \\
2 & $5280~(\times9.96)$~ & $2700~(\times5.09)$ & $\mathbf{530}$ \\
3 & $5040~(\times11.78)$ & $2695~(\times6.31)$ & $\mathbf{428}$ \\
\bottomrule
\multicolumn{4}{l}{(between parenthesis is the multiplication factor compared to our method.)}
\end{tabular}
\vspace{-2mm}
\end{table}
We also provide the training time per sequence in Table~\ref{tab:training_time}. Our method provides higher accuracy than MOISST, while requiring a much lower training time. The fact that MOISST with cropping performs much worse than without shows that it heavily relies on the RGB image correspondences to regularize the calibration, thus removing the top half of the images eliminates a lot of information for MOISST to exploit.
 With our progressive voxel resolution, the training time is 10 times shorter than MOISST, and 5 times shorter than MOISST with cropping.
\begin{figure*}[!htbp]
\centering
\scriptsize
\setlength{\tabcolsep}{0.002\linewidth}
    \begin{tabular}{cc}
\includegraphics[width=0.30\linewidth]{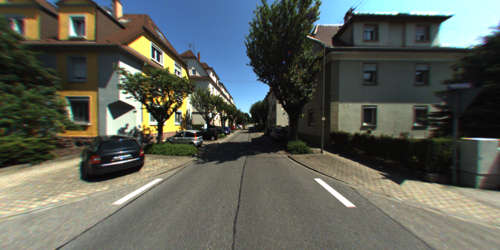} &
\includegraphics[width=0.30\linewidth]{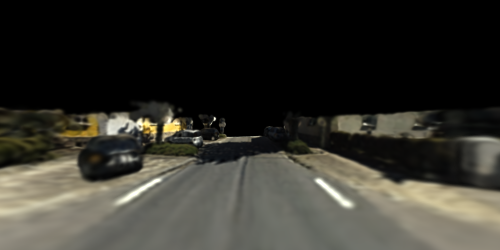}\\
Input RGB&10 cm\\
\includegraphics[width=0.30\linewidth]{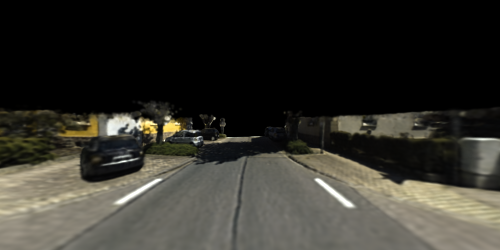}&
\includegraphics[width=0.30\linewidth]{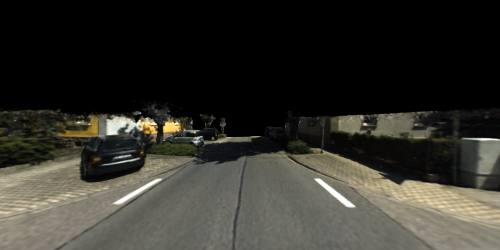}\\
5 cm&2 cm\\
    \end{tabular}
\caption{\textbf{Rendering results with different voxel sizes:} The finer details of the scene require a smaller voxel size to be learned.
}
\label{fig:voxel_size_image}
\vspace{-3mm}
\end{figure*}

\subsection{LiDAR/Camera calibration}
For the LiDAR/camera calibration task, we remove the temporal calibration component. 
We reduce the number of steps in our method to 3000 (2000 steps with 10 cm voxels, 500 with 5 cm, and 500 with 2 cm).
In addition to MOISST, using the code provided by the authors, we add the comparison to a classical method based on mutual information proposed by Pandey et al.~\cite{pandey2012automatic}, and a deep learning-based method, LCCNet~\cite{lv2021lccnet}. For LCCNet, We use the provided weights pre-trained on KITTI~\cite{geiger2013vision}, a dataset similar to KITTI-360~\cite{liao2022kitti}, albeit with some differences in the setup. 
\begin{figure}[!htbp]
\vspace{-3mm}
\centering
\scriptsize
\setlength{\tabcolsep}{0.002\linewidth}
    \begin{tabular}{ll}
          \rotatebox{90}{~~~~Pandey et al.\cite{pandey2012automatic}} & \includegraphics[width=0.95\linewidth]{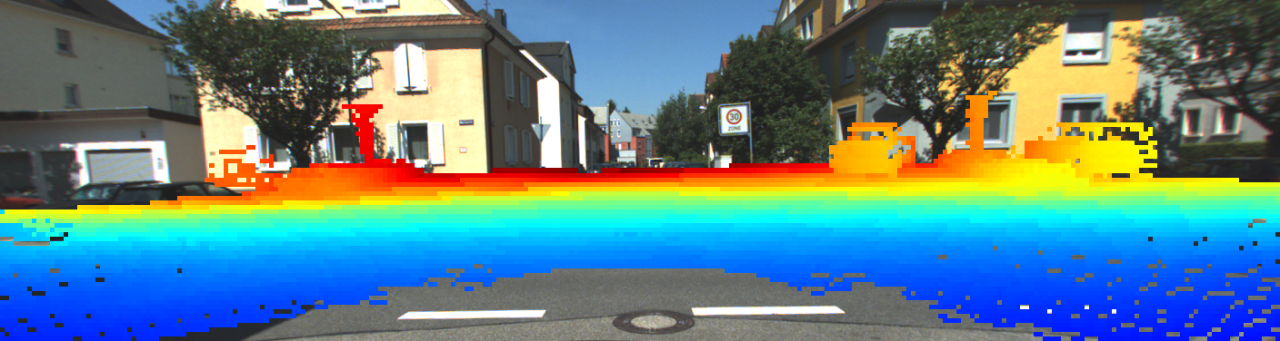}\\
          \rotatebox{90}{~~~~~~~~LCCNet\cite{lv2021lccnet}} & \includegraphics[width=0.95\linewidth]{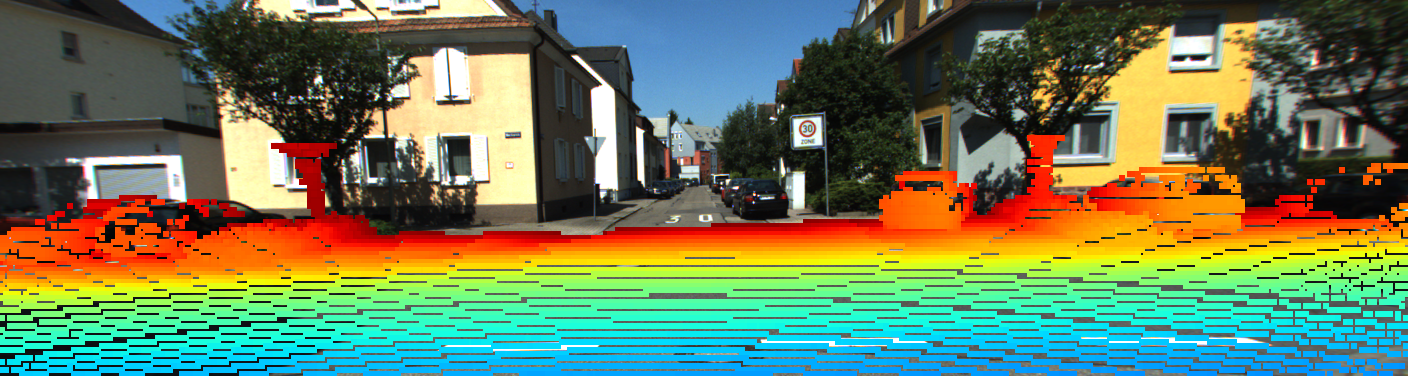}\\
          \rotatebox{90}{~~~~~MOISST~\cite{herau2023moisst}} & \includegraphics[width=0.95\linewidth]{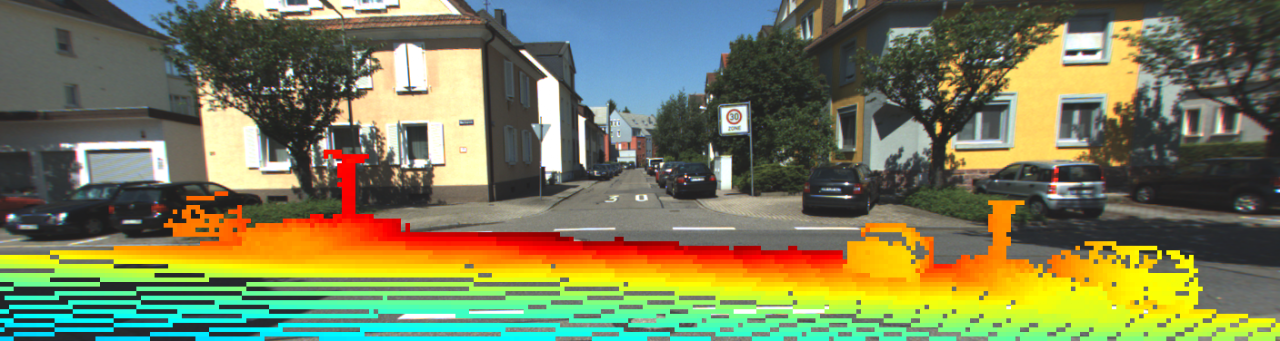}\\
          \rotatebox{90}{~~~3DGS-calib~(ours)} & \includegraphics[width=0.95\linewidth]{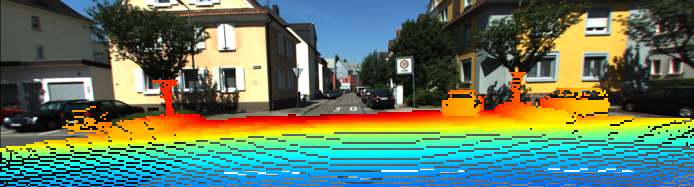}\\
    \end{tabular}
\caption{\textbf{LiDAR/Camera calibration results:} Point cloud to camera reprojection results obtained from the compared methods.}
\label{fig:lidar_reproj}
\vspace{-5mm}
\end{figure}
We evaluate this task using the front-left camera and the LiDAR of the sequences, and report the results in Table~\ref{tab:lidar_cam} and Figure~\ref{fig:lidar_reproj}. As shown by~\cite{herau2023soac}, LCCNet weights are system-specific, explaining the results as it was pre-trained on KITTI, where the camera/LiDAR placement is different.
Pandey et al.~\cite{pandey2012automatic} is unable to provide satisfying results since in-the-wild sequences in urban areas have a lot of shadows or elements where there is no correlation between grayscale image and LiDAR reflectance. MOISST, when not using a camera as a reference sensor, seems unable to provide satisfying calibration, as it was not able to correctly merge the LiDAR geometry with the one built from the images (see Figure~\ref{fig:moisst_lidar_fail}).

\begin{table}[hbt!]
\centering
\vspace{-2mm}
\caption{\label{tab:lidar_cam} LiDAR/Camera calibration}
\scriptsize
\setlength{\tabcolsep}{0.01\linewidth}
\renewcommand{\arraystretch}{1.1}
\vspace{-3mm}
\begin{tabular}{l | cc  cc }
    Method & Rotation (°) & Translation (cm) & Training time (s) \\
    \midrule
    Pandey et al.~\cite{pandey2012automatic} & $11.6 \pm 6.5$ & $90.7 \pm 29.5$  &  -\\
    LCCNet~\cite{lv2021lccnet} & $1.86 \pm 0.09$ & $89.5 \pm 2.2$  &  -\\
    MOISST~\cite{herau2023moisst} & $4.92 \pm 0.5$ & $64.1 \pm 6.8$ & $1527$\\
    MOISST w/ cropping & $6.01 \pm 1.2$ & $75.0 \pm 3.7$  & $860$\\
    3DGS-Calib (Ours) & $ \mathbf{0.45\pm 0.2}$ & $ \mathbf{9.60\pm2.1}$ & $\mathbf{216}$ \\
\bottomrule
\end{tabular}
\vspace{-3mm}
\end{table}
\begin{figure}[!htbp]
\centering
\scriptsize
\setlength{\tabcolsep}{0.002\linewidth}
    \begin{tabular}{ll}
            \rotatebox{90}{~~~~~Input RGB} & \includegraphics[width=0.95\linewidth]{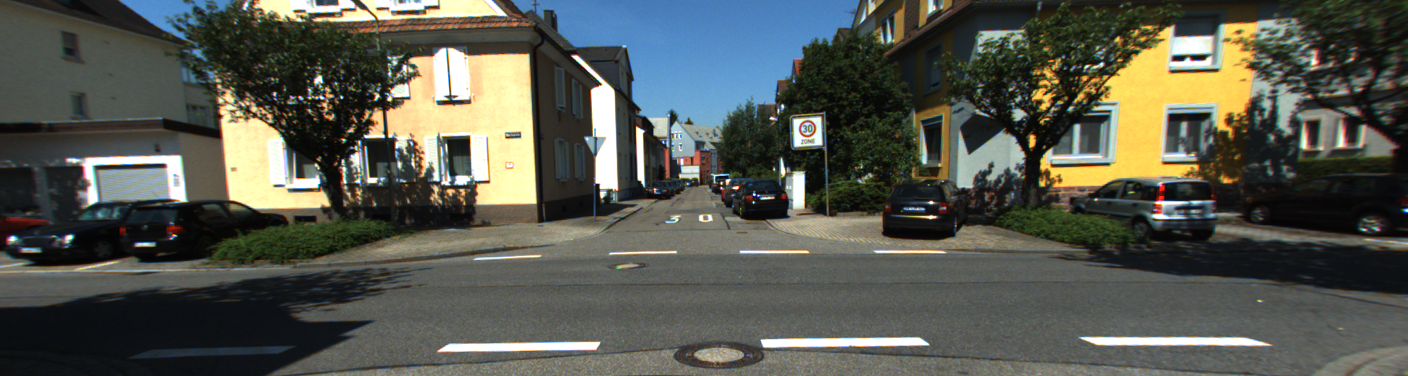}\\
            \rotatebox{90}{~~~~~MOISST RGB} & \includegraphics[width=0.95\linewidth]{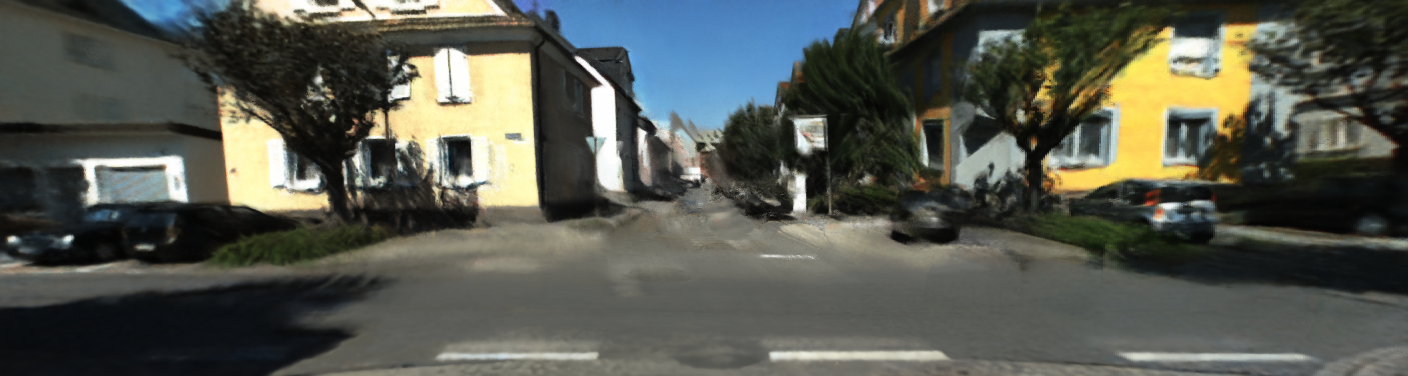}\\
            \rotatebox{90}{~~~~MOISST Depth} & \includegraphics[width=0.95\linewidth]{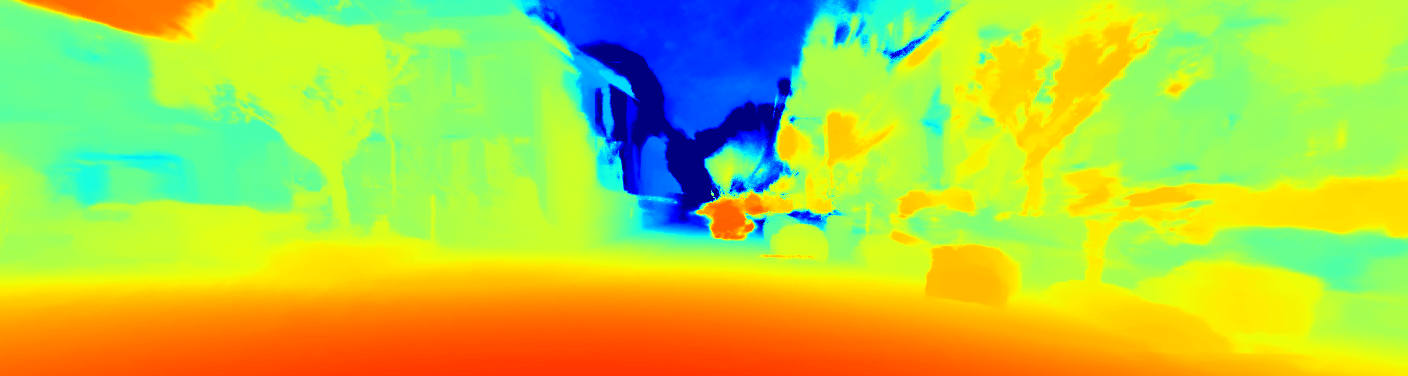}\\
    \end{tabular}
\caption{\textbf{MOISST rendering with LiDAR/Camera calibration:} MOISST fails to merge the geometry of the trees and cars from the LiDAR with the geometry built with the images.}
\label{fig:moisst_lidar_fail}
\vspace{-5mm}
\end{figure}

\subsection{Ablation study}
We compile the calibration results and training times according to the voxel size, as well as the number of Gaussians for each resolution in Table~\ref{tab:ablations} and the number of points for each resolution in Table~\ref{tab:nb_gaussian}. It can be noticed that a smaller voxel size allows to learn finer details (Cf. Figure~\ref{fig:voxel_size_image}), but requires much higher training time and more easily converges to local minimum. Our progressive voxel size reduction provides a balance between speed and performance. 
The rendering results for different voxel resolutions are provided in Figure~\ref{fig:voxel_size_image}.
\begin{table}[hbt!]
\centering
\scriptsize
\caption{\label{tab:nb_gaussian} Number of points for each sequence and voxel size.}

\renewcommand{\arraystretch}{1.2}
\begin{tabular}{@{}c c c c c c @{}}
Sequence & Original & 2cm & 5cm & 10cm \\
\toprule
1 & 4,068,707 & 3,289,412 & 1,295,133 & 394,154\\
2 & 4,025,853 & 3,351,941 & 1,431,667 & 484,690\\
3 & 4,220,417 & 2,948,428 & 903,399 & 245,549\\
\bottomrule
\end{tabular}
\vspace{-2mm}
\end{table}
\begin{table}[hbt!]
\centering
\scriptsize
\caption{\label{tab:ablations} Spatiotemporal calibration accuracy and training time for different voxel sizes.}

\setlength{\tabcolsep}{5pt}
\renewcommand{\arraystretch}{1.1}
\begin{tabular}{@{}c c c c c c c c@{}}

Voxel size & Rotation & Translation &Time offset & Training time \\
& (°) & (cm) &(ms) &(s) \\

\toprule
10 cm & $\mathbf{0.28}$ & $14.1$ & $8.3$ & $\mathbf{285}(\times0.59)$ \\
5 cm & $\mathbf{0.28}$ & $\underline{10.1}$ & $\underline{7.2}$ & $620(\times1.26)$ \\
2 cm & $0.38$ & $\mathbf{9.0}$ & $7.9$ & $1390(\times2.85)$ \\
Ours & $\underline{0.31}$ & $10.3$ & $\mathbf{6.7}$ & $\underline{490}~~~~~~~~~$ \\
\bottomrule
\multicolumn{5}{l}{(between parenthesis is the multiplication factor compared to our method.)}
\end{tabular}
\vspace{-5mm}
\end{table}

\section{CONCLUSIONS AND FUTURE WORK}
\label{sec:conclusions}
In this paper, we proposed a novel targetless multimodal and multi-sensor calibration method that takes advantage of the speed of 3D Gaussian splatting to provide an accurate and robust calibration with shorter training time compared to NeRF-based methods.
Our experimental evaluations on the KITTI-360 dataset demonstrate that we achieve better calibration results than both classical and recent NeRF-based methods while being orders of magnitude faster.
In the future, we expect to remove the assumption of the LiDAR points being concentrated in the bottom half of the images, to allow better adaptability to uncommon types of LiDAR or unusual camera dispositions.

\addtolength{\textheight}{-0cm}   






{\small
\bibliographystyle{IEEEtran}
\bibliography{cvpr_biblio}
}

\end{document}